# The Limits of Computation in Solving Equity Trade-Offs in Machine Learning and Justice System Risk Assessment

Jesse Russell[1] [0000-0002-4856-2462]

[1]Reno NV 89506
Jesse.r.russell@gmail.com

**Abstract.** This paper explores how different ideas of racial equity in machine learning, in justice settings in particular, can present trade-offs that are difficult to solve computationally. Machine learning is often used in justice settings to create risk assessments, which are used to determine interventions, resources, and punitive actions. Overall aspects and performance of these machine learning-based tools, such as distributions of scores, outcome rates by levels, and the frequency of false positives and true positives, can be problematic when examined by racial group. Models that produce different distributions of scores or produce a different relationship between level and outcome are problematic when those scores and levels are directly linked to the restriction of individual liberty and to the broader context of racial inequity. While computation can help highlight these aspects, data and computation are unlikely to solve them. This paper explores where values and mission might have to fill the spaces computation leaves.

## 1 Machine Learning Models in Justice Applications

This paper explores potential trade-offs that might occur when attempting to make machine learning models more racially equitable. It focuses specifically on inequity and its implication in justice settings. This paper is not a full exposition of different kinds of fairness. Also, it is not an investigation into different transformation methods that might be used to clean a data set of group difference (such as by gender or by racial groups). Instead, this is a consideration of some areas of machine learning in justice settings as risk assessments where either equity considerations and empirical considerations can be in tension with each other, or where different ethical considerations are in tension with each other. The purpose of the current paper is to highlight areas where computation might not be a solution or might not provide clear guidance on how these trade-offs might be made.

Many papers have catalogued different definitions of fairness that might be applied to machine learning. For example, Verma and Rubin [1] identify more than twenty different forms of equity that have been proposed for machine learning. Rather than discovering a common basis for all ethics in machine learning, they find that a single study might be found to be either fair or unfair depending on the fairness measures applied.

Similarly, Corbett-Davies and Goel [2] identify three prominent measures of equity. These are a) protected attributes, like race, are not included in the model; b) performance, such as false positives, is equal across groups; and c) outcome rates by calculated risk are similar for different groups). Though prominent measures, Corbett-Davies and Goel find that these fairness concepts suffer from significant statistical limitations. While each of these three has appeal, it is not always possible to maximize all three simultaneously. In fact, the authors state that applying one of these fairness measures could make fairness of the model worse, not better.

Similarly, Berk et al. [3] point out the main tension with fairness measures – they often come at the cost of performance or accuracy. If the most accurate model is unfair, then anything that improves fairness will detract from accuracy. Berk, et al., find that "Except in trivial cases, it is impossible to maximize accuracy and fairness at the same time, and impossible simultaneously to satisfy all kinds of fairness" (p. 1, 2018). Fairness is often more normative than computational. There is no single measure of fairness that, once reached, a model can be said to have achieved fairness. Fairness as a normative value has to be negotiated with computational value and with business value in how it is applied in practice.

The question of fairness, particularly racial equity, is especially important in a justice context. The research literature has found that the justice system is inequitable over several dimensions. One study found that 60 percent of incarcerated people in the United States were Black and Hispanic [4]. Another found that Black people are 83 percent more likely to be arrested for marijuana compared to White people, despite similar rates of marijuana usage across racial groups [5]. Another study of 60 million state patrol stops found that Black drivers were stopped more often and the reasons for stopping Black drivers were less serious than the reasons for stopping White drivers [6]. Another study of youth used propensity scoring to conclude that race influenced the likelihood of detention intake, adjudication, and disposition, with Black youth receiving harsher treatment at each stage [7]. Another study of youth found that Black males received a higher likelihood of a formal adjudication, even after controlling for legal history and the details of the current offense. One group of researchers commented that, "The large racial disparity in incarceration is striking" [8]

The consequences of justice system involvement or penetration are markedly negative. One study [9] found people released from incarceration struggle in the labor market after their period of incarceration. Empirical results show that in the first full calendar year after release, only 55 percent have any reported earnings at all. Incarceration has also been shown [10] to raise the risk of divorce and separation and shown to reduce family financial resources. Research has indicated that incarceration leads to poorer health outcomes, financial strife, and impaired social standing [11].

Empirical research has found that for young people, any contact with the juvenile justice system can have negative impacts and cause twice the likelihood of being arrested as adults, compared to similar youth with similar behaviors who did not have contact with the juvenile justice system. The study found that controlling for other factors, incarcerated youth were 38 times as likely as other youth with equivalent backgrounds and self-reported offending histories to be sanctioned for crimes they committed as adults [12]. Another study [13] found that juvenile justice out of home placement was negatively related to the likelihood of high school graduation. They found that youth, once incarcerated, were unlikely to ever return to school. The study also found that juvenile incarceration increased the likelihood of adult incarceration. A study of 40,000 Florida youth found that those assessed as low risk and placed into a residential facility were rearrested more often than similar youth who were not placed and also more often than high risk youth who were not placed [14].

The potential for tension between data-driven accuracy and normative fairness can be magnified in a justice context where the consequences of accurate public safety decisions and fairness with a race-equity lens are especially momentous. This was highlighted by Corbett-Davies, et al. [15] in the context of justice and public safety. In a context in which the identification of true positives and the avoidance of false negatives can have a direct public safety consequence, they find that models that are more accurate (and thus promote public safety) are in tension with fairness. Changing the goals of machine learning or of justice risk modeling away from predictive accuracy to a more inclusive set of goals around fairness, leads to less accurate models and thus less public safety.

## 2      Nine Points of Tension

In this context, this paper considers nine points of tension that might occur in using machine learning to develop a risk assessment in a justice context. These points of tension do not offer easy solutions. They are not Gordian Knots waiting for a sword. Rather they are instances where data scientists must negotiate between the data and the computational space and the normative and values space. Some of these points of tension rest in demographic differences among those appearing the datasets used for developing a risk assessment using machine learning; others may be shaped by calibration choices made for the machine learning model, like where cut points are set or how class weights are set; and others are more about differences in rates of interventions rather than differences in statistical odds themselves, like differences in the likelihood of a probation sentence versus a confinement sentence. Further, while each of these points of tension represent something different, they often intersect with each other in practice. Some problems, like different base rates across racial groups of the outcome occurring can directly impact other things, like different likelihoods of getting a high score or a different likelihood of experiencing the outcome given a high score. It is precisely because these points of tension relate to each other that they are difficult to solve.

In justice systems, "risk assessments" are used to inform a variety of decision points. Which risk assessment tools and which decision points vary widely across jurisdiction, but some commonalities exist. Most risk assessments are applied at a current legal decision point, use information about the person's legal history along with other information gathered, and produce a predictive classification of a future legal event. The future legal event might be something like being charged with a new law violation, or something requiring a legal decision, like an arraignment or adjudication, or even disposition or sentencing. In other cases, the future legal event is not based on a new crime, but rather is based on compliance, like obeying probation rules, or for youths can include things like obeying parents. The time frame for these outcomes to occur also vary. They might be as short as a few days while waiting for a judge to make a ruling or could be over the course of two years while desistence is tracked.

Even with all the permutations of risk assessments they mostly operate in the same way. Some set of features are used to create a risk score and a risk level, and the score and the level shape the decisions of the justice system for the person being scored to either increase or decrease the person's penetration into the justice system.

### 2.1    Similar likelihoods of getting any particular score? [11]

Should there be an expectation of similar likelihoods? Should it be a measure of risk model performance that it produces even score distributions (see Baird, et al. for a discussion of distributions as a measure [16]) across racial groups? Or should risk models be agnostic about likelihood differences if they are an attribute of the data? If a data scientist finds that the machine learning model is most accurate when there are large differences in score distributions across racial groups, should those differences be embraced? Or is it incumbent on model development to potentially sacrifice some accuracy in order to achieve more similar distributions? There is no computational solution to this question.

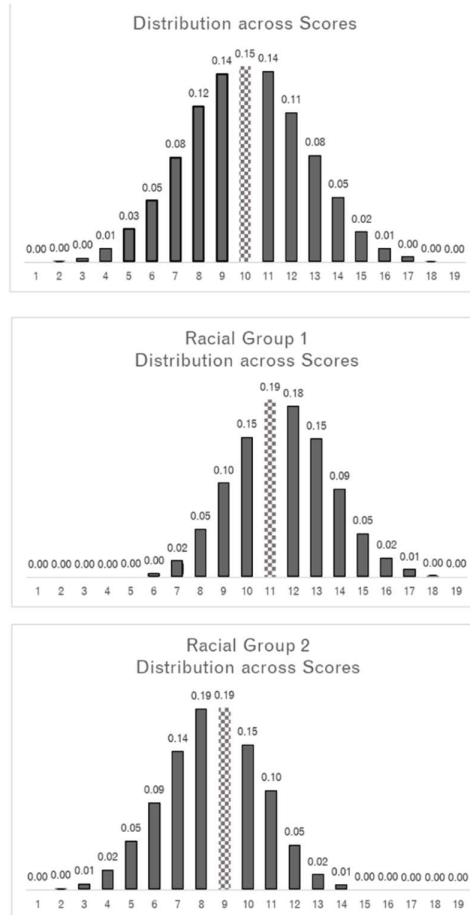

**Fig. 1.** Overall likelihoods of receiving a particular score might mask substantial differences by race. The example distribution in this graph shows that skews in opposite directions for racial groups can create overall distribution with no skew. Trying to un-skew either racial group could lead the overall distribution to be skewed.

### 2.2 Similar distribution of risk levels for each racial group? [12]

Even if the distribution across levels looks good overall, there can be huge differences in the distribution for each racial group. (See the COMPAS validation report for how distributions of risk levels care used [17].) In some ways this is the same conundrum as with scores, but there is a difference: different risk levels are directly tied to interventions (some supportive and some punitive). Where to establish cut points to separate the scores into something like high, moderate, and low can be informed by data, but cut points are more a facet of mission and values than of computation. Different cut points might exacerbate the inequality in scores by creating groups that have a high level of racial homogeneity. Cut points that might create something like three even groups might mask differences by race. Or cut points based on maximizing outcome rates across the levels might also put more of one racial group into one level and more of another racial group into another. Looking at receiver operating characteristic statistics or F1 score might be useful for making cut points that help balance a confusion matrix, but they might not help make cut points that are free from racial imbalance.

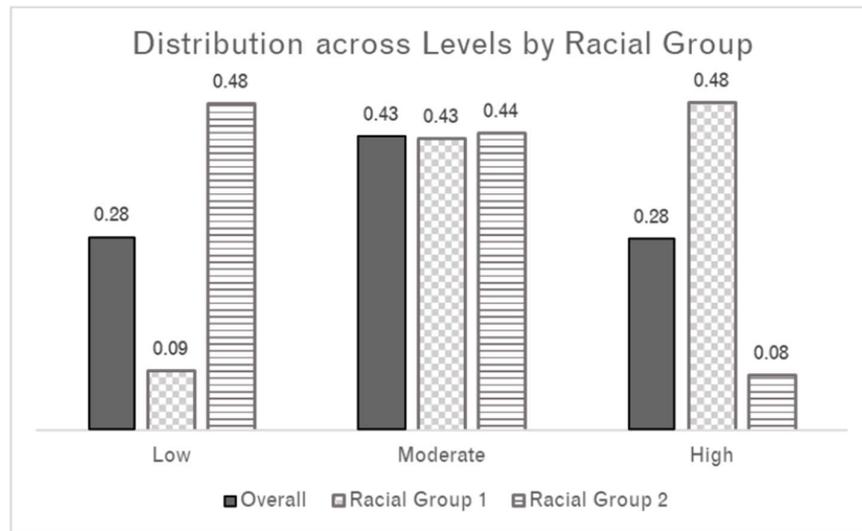

**Fig. 2.** Overall likelihoods of receiving a particular level can appear to be relatively normal while different racial groups have non-normal distributions. The example in this graph shows one racial group with very few low scores and the other racial group with very few high level scores. Together, they represent a distribution with an even frequency of low and high levels scores.

### 2.3 Similar likelihood of getting a high-risk score for each racial group?

What about similar likelihoods of getting classified as "high" which may mean out of home placement? (See Orbis Partners 2007 for an example of how a validation study considered proportions of each racial group classified into low, moderate, and high risk groups [18]). Looking at the proportion of all high-risk scores that are allocated to each racial group, it might be a goal to have this proportion match the proportion of different racial groups in the underlying population. But that might not be possible given the distribution of risk scores. If one group is disproportionately (compared to the population) assigned high risk scores, what are the consequences of practice? In many justice contexts, a high-risk label means out of home placement (such as detention, jail, prison). Can cut points be assigned without a consideration of the practice consequences of how high-risk scores and out of home placements are distributed? This question has to be solved through goals and objectives; it cannot be solved computationally.

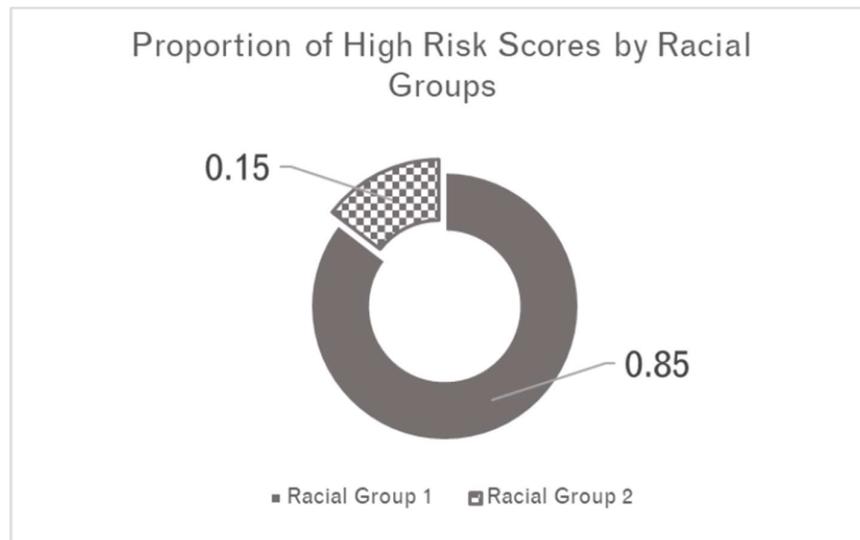

**Fig. 3.** Another perspective on the example distribution shown in Fig. 2 shows the proportion of all high risk scores going to one racial group. It is not only the conditional likelihood of a high score given that a person is a member of a particular racial group, it is the reciprocal conditional likelihood that a person is a member of the racial group given that they have a high score that matters.

### 2.4  Similar likelihoods of re-arrest across scores for each racial group?

Alternatively, equity could be framed not in terms of the distribution of scores or levels across racial groups but in terms of the likelihood of re-arrest across scores. (See the validation of the Y-LSI risk assessment for how re-arrest rates correspond to scores in similar and different ways across racial groups. [19]) The outcome rate could increase steadily across scores – meaning low scores were less likely to correspond to the outcome occurring and high scores corresponded more often to the outcome occurring. These stepwise increases might be an indicator that the risk model is functioning as it should: scores being a valid indicator of the outcome. But the overall distribution of the likelihood of the re-arrest outcome across scores might be distinctly different than the distribution for each racial group. In particular, the distribution for one group might be higher across all risk scores while the distribution for another group might be lower across all risk scores. This is common when the overall re-arrest rate is higher for one group than it is for another. In this case, the question is: should there be an expectation that every person have a similar chance of a re-arrest at a particular risk score regardless of race? Is it a problem if a risk score means something different (like a higher likelihood of the outcome occurring) for one group than it does for another (like a lower likelihood of the outcome occurring)? Even if it were a goal, with different re-arrest rates occurring in the community, there may be no way to change the machine learning model in a way that creates a similar distribution for each racial group. The machine learning algorithm learns that re-arrest rates are higher in the community for one group and reflects that fact in the result. But just because it is the statistical result, does not mean it upholds values around equity.

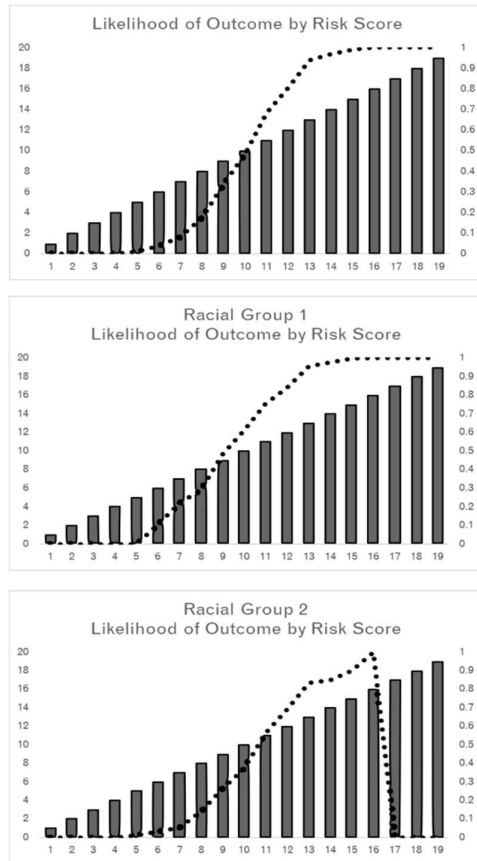

**Fig. 4.** Again, overall distributions, this time representing the likelihood of experiencing the outcome given a particular risk score, can obscure differences between racial groups. In the example distribution in this graph, it is clear that the overall distribution does not highlight the fact that one racial group experiences the outcome at a very low rate with high risk scores while the other racial group experiences the outcome much more often.

## 2.5 Similar likelihood of experiencing outcome by risk level for each racial group?

A similar dynamic occurs with outcome rates by risk level. (See the validation study of the Riverside Pretrial Risk Assessment Tool for how validation efforts can examine risk level distributions by race. [20]) They may look like a reasonable distribution overall, but clearly dissimilar outcome rates by level for each racial group might be present. Again, the question is a normative one: Should there be an expectation of similar likelihoods of the re-arrest outcome for each racial group across each risk level? Should what a risk level represents (lower, moderate, and higher likelihoods of the outcome occurring) be the same for every person, regardless of race? Should the label "high risk" consistently imply the same risk, regardless of race?

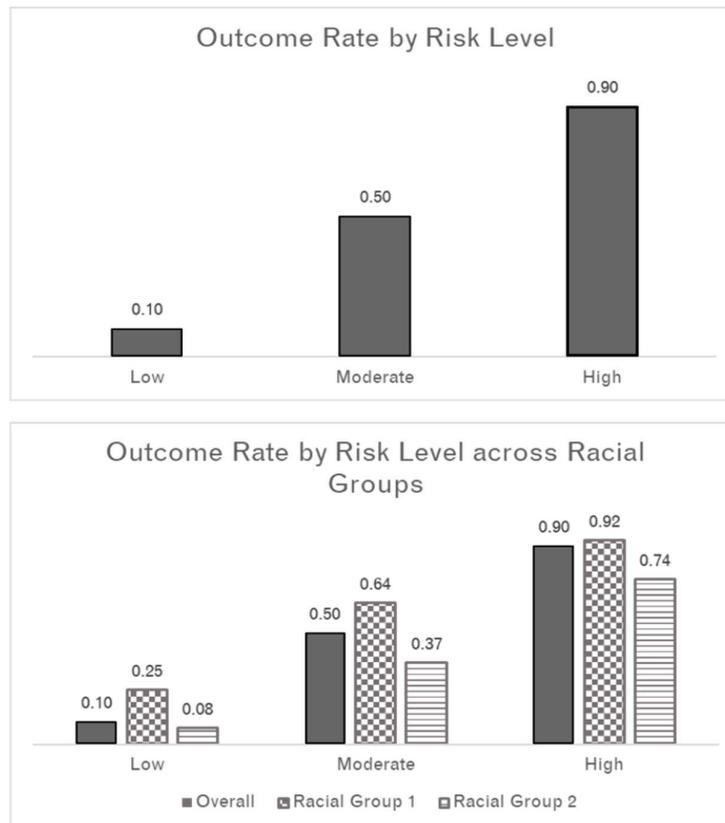

**Fig. 5.** In these graphs, the disjuncture between the overall rates and the racial group rates is marked. A step-wise increase in outcomes by risk level appear to offer good discrimination and separation by level. For practical purposes, those with high risk scores experience the outcome much more often than those with moderate risk scores. And those with low risk scores experience the outcome less often than those with moderate risk scores. However, the step-wise increase over the levels is much steeper for one group than the other. The outcome rate at high risk for one group is similar to the outcome rate at moderate risk for the other group.

### 2.6  Similar distributions of scores versus similar outcome rates by score?

A different question asks about characteristics of the distributions across scores by race as a tradeoff for characteristics of the likelihood of the outcome occurring by risk score and by race. There is generally a tradeoff where a more similar distribution of scores might mean a greater difference in outcome rates by level for different racial groups. Google Research provides a good tool for demonstrating the tradeoff at https://research.google.com/bigpicture/attacking-discrimination-in-ml/. Or more similar outcome rates by score might go with greater differences in the distribution of scores. How should it be decided which is more important: an even likelihood of getting a particular score, or an even likelihood of a re-arrest for a particular score? When the base rate of experiencing the outcome is different across racial groups, this tradeoff may be intractable. In a justice environment, this tradeoff can take on a different significance. In a context in which one group has experienced a long history of over-policing and over-incarceration, and where one group faces structural and institutional biases across the life course, and one in which negative justice outcomes are experienced more often by one group, is fairness achieved by saying that high risk relates to similar outcome rates across racial groups?

### 2.7  AUC and false positives as measures of performance across racial groups?

A different consideration for equity might consider whether the risk model performs similarly for each racial group. (For example, the validation study of Kentucky's Public Safety Assessment considers AUCs overall and by race. [21]) The area under the curve (AUC) overall and the AUC by racial group might look even, but the risk model might produce more of one kind of error for one group while the other has more of a different kind of error [22]. In particular, even with similar AUCs, one group might experience more false positives than another group. False positives in this context can be enormously negative. Over-intervening for an arrested youth, especially punitive interventions and ones that restrict a person's liberty, can have devastating outcomes. Mistakenly sending a youth to an out of home placement, a youth who was actually unlikely to be re-arrested, can severely disrupt their development, life path, family, and community. Trying to create a risk model that produces similar false positive rates can be difficult when initial referral rates are different across racial groups and re-arrest rates in the community are different. Even if the data the machine learning is based on minimizes errors in a way that creates an imbalance in false positive rates, are communities ready to embrace a risk tool that mistakenly sends youths to out of home placement for one race more often than for another?

On the other hand, false positives might be made to be equal across groups, but only by sacrificing overall performance, or by significantly skewing the distribution across scores and levels for each racial group.

The question might not even be about *equal* false positive rates. False positives might not just need to be equal, but in a world of structural inequality in justice system involvement, false positives might actually be a greater cost for one group than for a more privileged group. In a justice system that can be defined by its structural inequality and over representation of Black/African American youth, how can a false positive be quantified?

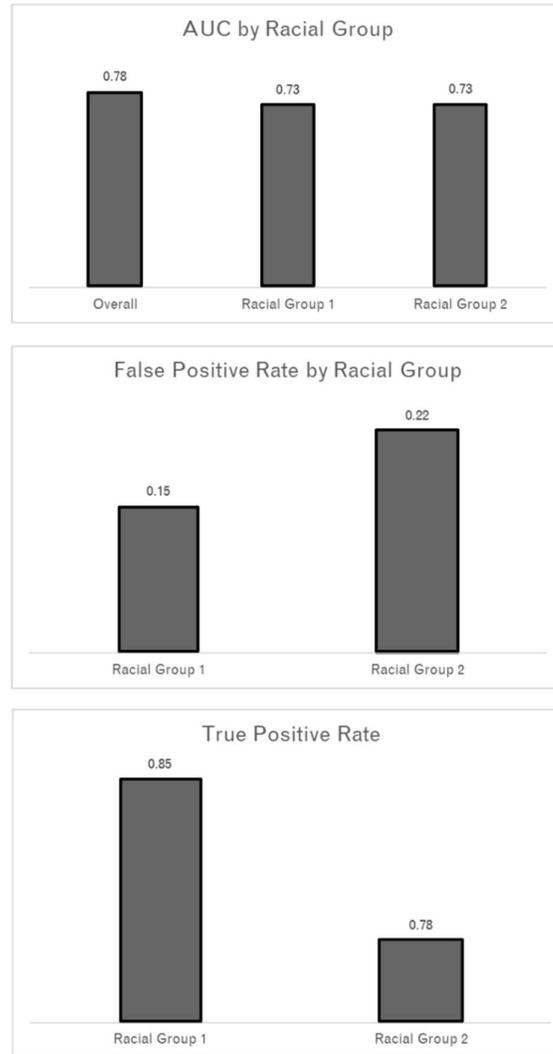

**Fig. 6.** Overall accuracy, measured by a common metric like the AUC, can be similar for different groups while underlying statistics are different for different racial groups. For the example measures in this graph, it can be seen that while AUC scores are similar overall, only looking at the aggregate accuracy score misses important difference in the kinds of errors each group experiences. Here, one group experiences many more false positives than false negatives while the other experiences many more false negatives than false positives. In a justice context, either false positives or false negatives might be considered more costly to system goals. And for individuals in the justice system, experiencing a false negative is much different from experiencing a false positive.

**2.8   Feature selection and structural inequality?**

Part of the underlying structure of machine learning in a justice context is that machine learning for feature selection for creating a risk model often selects risk items that correlate both with outcome and with race. For example, the Missouri Juvenile Risk Assessment includes the item "Parental History of Incarceration." [23] Since incarceration is experienced at higher rates by Black people than for White people, this item is likely to correlate with race. Even if it is a useful feature in the predictive model, it is probably a proxy indicator of race. This item acts to port structural racial bias in incarceration into how youths are scored into low, moderate, and high risk.

Or, a youth's legal history (maybe counted as the number of past referrals, or the number of past charges, or number of past adjudications) might be a useful feature for the risk model, but it is also likely related to race in many communities. Arrest rates in some communities are connected to structural inequalities. Things like housing, economic development, health, schools, or policing and patrol patterns can all be part of a larger set of structural inequalities that influence who is most likely to be arrested. Given those structural factors, it is common for Black/African American youth to be re-arrested more than others. In this case, features that help a risk model's accuracy can inadvertently be bringing those structural inequalities into the risk model. Even without implicit bias or intentional racism, the factors chosen for the risk model can bring prejudice bias into the model.

This presents a puzzle for the data scientist. One option is to try to strip the features of their racial relationships. Different procedures exist to try to create unbiased proxy variables. But in a context in which all potential variables have a racial element, these procedures do not solve the underlying problem. Another option is to search for different features that are not correlated with race. Again, this can be difficult or impossible. Since the likelihood of the outcome is correlated with race, anything correlated with the likelihood of the outcome occurring is also going to be correlated with race. Even finding features that have lower correlations with race might be difficult, and they might also be less predictive of the outcome. It is indefensible to tell an impacted community that a risk tool must be unfair because structural inequality in the community says it must be.

**2.9   Intersection of "high risk and race in out of home placement decision making**

Risk assessments in justice practice generally help inform decision-making while leaving wide latitude for discretion in place. Probation officers, prosecutors, court staff, and judges often use risk assessment outputs as one piece of information among others. While often matching the realities of practice and even of the law, it does allow for implicit bias to become part of the decision making. In particular, the "high risk" designation might have different impacts for different racial groups. Research has shown that Black/African American youth are seen as more threatening, larger, and older. And Black/African American youth might come into the justice system with fewer supports [24]. In practice, this means that discretionary decisions about out-of-home placement for youth with high risk scores, might lead to a higher proportion of Black/African American youth being sent to a facility. No matter how data-driven the model is, how thorough the data science, how insightful the machine learning is, the model can intersect with bias when it is applied in practice. This is especially a problem given the negative outcomes associated with out of home placement.

## 3   Conclusion

In the end, there are three facts in the community that are transmitted through the data and into the risk mode. One is that different racial groups experience a given outcome (such as re-arrest) significantly more or less often than others. The second is that features that are useful for predictive accuracy tend to be correlated with one racial group. And the third is that these differences (in outcomes and in feature relationships) are related to historical and structural inequalities in the community. In this context, there can be no computational solution to inequality. Any solution will have to be based in the normative values of impacted communities, stakeholders, people who are part of the justice system, and leadership. Further, there is likely no one solution to the tangle of competing values. Maximizing any one dimension of fairness will have to come at the sacrifice of another aspect of fairness and at a cost to accuracy. In a context in which lack of fairness and accuracy have potentially disastrous outcomes for individuals and communities, these trade-offs must belong to a wide coalition of impacted people, communities, stakeholders, people from the justice system (such as probation officers), the courts (such as prosecutors or judges), and from administrative, political, and community leadership.

In the justice environment, machine learning and machine learning based risk assessments are an increasing point of conflict. Machine learning and machine learning risk assessments often belong to the vendors who develop them. And these vendors sometimes avoid the topic of racial bias entirely. Some, like Equivalent (previously Northpointe) do not share any details of how their tools are created, validated, or relate to racial bias. And when others do publish results of predictive validity testing, there is often no consideration of race. [25]

The critiques of these tools often come from the people and communities most impacted by them, community advocates, and academics. These people and communities see machine learning and machine learning based risk assessments as an embodiment of structural racism that should be removed from the justice system. Commentary is often absolutist, like the statement that "there is no way to develop a system that can predict or identify "criminality" that is not racially biased — because the category of 'criminality' itself is racially biased." [26]

While there is no agreed upon perspective on machine learning fairness in justice settings, a reconciliation is unlikely. As long as there is no clear space where both structural racism and machine learning can exist, machine learning in justice systems will be problematic. As some look to computation and better machine learning models as the solution to inequity in justice systems, others see the use of data, computation, and machine learning efforts as a rationalization and a veil for inequity.

The solution to the tensions might have as much to do with how machine learning models are used in practice, as it does the development and validation of machine learning tools. Inasmuch as machine learning risk assessments are used to target punishments, to assign negative consequences, and to restrict the liberty of people in the system, then risk assessments will continue to face these tensions. But if risk assessments were used to allocate resources, support positive development for people in the system, and to elevate community partners and interventions, then much of the tension is released. In this way, machine learning in justice contexts could potentially shift from anticlassification to antisubordination, though this would represent a major practice change for most justice agencies. [27]

Further, if the factors that predict future outcomes like re-arrest, are seen being borne solely by the individual being assessed, then the model misses the ecological aspects of what drives justice system involvement. A consideration of things like a history of housing discrimination, underperforming schools, and high levels of involvement in child welfare might be missed when the models are focused on individual behaviors. As Barabas, et al, [28] suggest, the underlying drivers of crime might be systemic as much as individualistic. An ecological view of justice system involvement might be the best way for machine learning to be a part of the justice system ethics conversation.